\pdfoutput=1

\documentclass[11pt]{article}

\usepackage[preprint]{acl}

\usepackage{times}
\usepackage{latexsym}
\usepackage{mdframed}  

\usepackage[T1]{fontenc}

\usepackage[utf8]{inputenc}

\usepackage{microtype}

\usepackage{inconsolata}

\usepackage{graphicx}
\usepackage{subfig} 

\usepackage{tabularx} 
\usepackage{booktabs} 
\usepackage{array} 
\usepackage{amssymb} 
\usepackage{pifont} 
\newcolumntype{P}[1]{>{\centering\arraybackslash}p{#1}} 

\usepackage[most]{tcolorbox} 
\tcbuselibrary{breakable}

\title{Towards Reliable Retrieval in RAG Systems for Large Legal Datasets}

\author{
  \textbf{Markus Reuter\textsuperscript{1,\footnotemark[1],\footnotemark[2], \footnotemark[3]}},
  \textbf{Tobias Lingenberg\textsuperscript{1,\footnotemark[1],\footnotemark[2]}},
  \textbf{Rūta Liepiņa\textsuperscript{2}},
  \textbf{Francesca Lagioia\textsuperscript{3,5}},\\
  \textbf{Marco Lippi\textsuperscript{2}},
  \textbf{Giovanni Sartor\textsuperscript{3,5}},
  \textbf{Andrea Passerini\textsuperscript{4}},
  \textbf{Burcu Sayin\textsuperscript{4}}
  \\
  \\
  \textsuperscript{1}{Department of Computer Science, Technical University of Darmstadt} \quad
  \\
  \textsuperscript{2}Department of Computer Science, University of Florence, \texttt{name.surname@unifi.it} \quad
  \\
  \textsuperscript{3}ALMA-AI, Faculty of Law, University of Bologna,  \texttt{name.surname@unibo.it} \quad
  \\
  \textsuperscript{4}DISI, University of Trento, \texttt{name.surname@unitn.it} \quad
  \\
    \textsuperscript{5} Department of Law, European University Institute \quad
    \\
}

\begin{document}
\maketitle

\begingroup
\renewcommand\thefootnote{\fnsymbol{footnote}}
\footnotetext[1]{Equal contribution.}
\footnotetext[2]{Work was performed at University of Trento.}
\footnotetext[3]{Correspondence: \href{mailto:markus.reuter@stud.tu-darmstadt.de}{markus.reuter@stud.tu-darmstadt.de}}
\endgroup

\begin{abstract}
Retrieval-Augmented Generation (RAG) is a promising approach to mitigate hallucinations in Large Language Models (LLMs) for legal applications, but its reliability is critically dependent on the accuracy of the retrieval step. This is particularly challenging in the legal domain, where large databases of structurally similar documents often cause retrieval systems to fail. In this paper, we address this challenge by first identifying and quantifying a critical failure mode we term Document-Level Retrieval Mismatch (DRM), where the retriever selects information from entirely incorrect source documents. To mitigate DRM, we investigate a simple and computationally efficient technique which we refer to as Summary-Augmented Chunking (SAC). This method enhances each text chunk with a document-level synthetic summary, thereby injecting crucial global context that would otherwise be lost during a standard chunking process. Our experiments on a diverse set of legal information retrieval tasks show that SAC greatly reduces DRM and, consequently, also improves text-level retrieval precision and recall. Interestingly, we find that a generic summarization strategy outperforms an approach that incorporates legal expert domain knowledge to target specific legal elements. Our work provides evidence that this practical, scalable, and easily integrable technique enhances the reliability of RAG systems when applied to large-scale legal document datasets.\footnote{Code is available at \url{https://github.com/DevelopedByMarkus/summary-augmented-chunking.git}.}
\end{abstract}

\section{Introduction}

Large Language Models (LLMs) are increasingly adopted in high-stakes domains such as law. Yet, they remain critically limited by the phenomenon of \emph{hallucination}: incorrect outputs that are fabricated or deviate from the provided source material, posing severe risks in legal applications \cite{huang2025survey, li2023dark, qin2024exploring}. Recent studies report hallucination rates between 58–80\% for general-purpose LLMs on legal tasks \cite{dahl2024large}, highlighting how factual reliability is not just desirable but essential for deploying LLMs in the legal domain.

This challenge is amplified by forward-looking proposals for how legal documentation itself may evolve. For instance, \citet{palka2025make} suggest that privacy policies may intentionally become longer and more comprehensive to ensure they are legally complete. In such a future, LLMs are expected to serve as the designated “readers” of these texts, extracting and summarizing information for human users. This vision, however, can only be realized if the systems are \emph{highly reliable}.

Retrieval-Augmented Generation (RAG) \cite{lewis2020retrieval} has emerged as one of the leading approaches to improving reliability. Using a trusted text corpus to provide factual evidence, RAG guides the LLM's output, reducing hallucinations and ensuring closer alignment with the source material \cite{tonmoy2024comprehensive}. In the context of long, structurally similar legal documents, identifying the relevant text passage as “needle in the haystack” becomes a top priority that we aim to address.

On a technical level, we quantify the retrieval quality with our Document-Level Retrieval Mismatch (DRM) metric and the character-level precision and recall. Then, we investigate a simple yet effective technique to improve retrieval quality, Summary-Augmented Chunking (SAC). We enrich text chunks in the trusted text corpus with document-level summaries. This preserves global context, lost in standard chunking, guiding the retriever toward the correct document without altering the underlying retrieval pipeline. This method is applied to question-answering tasks across a diverse set of legal documents, including privacy policies, non-disclosure agreements, and merger-and-acquisition contracts.

\textbf{Key contributions:} 
\textbf{(i)} First, we define and quantify \emph{Document-Level Retrieval Mismatch} (DRM), a key failure mode we observe in standard RAG pipelines where the retrieved information originates from the entirely wrong source document.
\textbf{(ii)} We propose \emph{Summary-Augmented Chunking} (SAC) as a lightweight and modular solution that strongly reduces DRM by injecting global context directly into each chunk. We experimentally validate SAC on Legalbench-RAG \cite{pipitone2024legalbench}, showing substantial improvements over standard chunking.
\textbf{(iii)} Additionally, we explore how to utilize legal domain knowledge by evaluating both a generic and an expert-guided summarization strategy. Interestingly, we find that simple, general-purpose summarization yield the best retrieval performance.

\section{Background and Related Work}
\subsection{Retrieval-Augmented Generation}

RAG \cite{lewis2020retrieval} is a powerful paradigm that enhances the reliability of LLMs by grounding their outputs in external knowledge sources. This approach is particularly crucial in high-stakes domains like law, where factual accuracy is not just desirable but mandatory. The standard RAG pipeline consists of two main stages: a retriever that searches a large document corpus to find text snippets or ``chunks'' that are relevant to a user's query and a generator model that synthesizes a final answer based on these retrieved chunks.

RAG has been adapted for a wide array of legal tasks \cite{hindi2025enhancing}, including case reasoning \cite{yang2024casegpt}, legal judgment prediction \cite{peng2024athena}, and legal question-answering \cite{cherubini2024improving, visciarelli2024savia}. Recent approaches have focused on improving reliability by imposing more structure on the knowledge source, for instance, through knowledge graphs \cite{kalra2024hypa} or structured case databases \cite{wiratunga2024cbr,jayawardena2024scale}.

Despite its demonstrated ability to improve factual accuracy in context-sensitive tasks \cite{gupta2024comprehensive}, the effectiveness of RAG is critically dependent on the quality of the initial retrieval step \cite{huang2025survey, hou2024gaps, hou2024clerc}. 
If the retriever fails to select information that is relevant or complete, the generator may produce factually unsupported responses. Accordingly, prior studies demonstrate that RAG is not a guaranteed solution, as legal RAG systems continue to generate a considerable amount of hallucinated content, particularly when the retrieval mechanism is ineffective \cite{dahl2024large, magesh2025hallucination,ariai2024natural}. 
Our work addresses this challenge by focusing on the \emph{pre-retrieval stage}, the engineering of the knowledge base, which forms the foundation of any reliable RAG system.

\subsection{Unique Challenges of Legal Text for RAG}

Legal documents present a major challenge for automated text processing systems due to their specific linguistic and structural characteristics \cite{ashley2018automatically, ferraris2024legal, liepina2019gdpr, martinelli2023ai}. These challenges make retrieval particularly prone to errors.

\textbf{(i) Lexical Redundancy:} Legal language is highly standardized, featuring boilerplate clauses, formally defined phrases, and specialized terminology that are often repeated across thousands of documents \cite{akter2025comprehensive}. For example, Non-Disclosure Agreements within a database may be structurally almost identical, differing only in a few critical variables such as party names or dates. This high degree of similarity can easily confuse retrieval models that rely on surface-level keyword matching or vector similarity \cite{joshua2025domain}.

\textbf{(ii) Hierarchical Structure:} Legal texts are organized in complex layouts with nested sections, subsections, and dense cross-references. Standard chunking strategies ignore document hierarchy \cite{ferraris2024legal, zilli2025agentic}, which cuts off these logical connections. As a result, retrieved chunks may appear relevant but lose their intended meaning when disconnected from their structural context.

\textbf{(iii) Fragmented Information:} Answering a legal question often requires synthesizing information scattered across multiple sections or even different documents \cite{hindi2025enhancing}. For example, interpreting an exception clause in a privacy policy may depend on definitions or stipulations introduced much earlier in the document. Retrieval systems must therefore go beyond finding locally relevant chunks and instead capture distributed factual dependencies that contribute to a legally meaningful answer \cite{bendahman2025eur}.

\textbf{(iv) Provenance and Traceability:} In legal applications, the provenance of information is of high importance. Answering a question correctly is insufficient if the supporting text is retrieved from the wrong source document \cite{uke2025generative}. For instance, pulling a clause from a similar but distinct contract would undermine the legal validity of the generated output and erode user trust \cite{joshua2025domain, hindi2025enhancing}. Consequently, legal professionals require a transparent and verifiable “reasoning trail” from the generated answer back to the specific clauses in the original source document \cite{richmond2024explainable}. This need for an auditable path, where every piece of information can be validated against its source, makes document-faithful retrieval a fundamental measure of a system's reliability.

\begin{figure*}
    \centering
    \includegraphics[width=1\linewidth]{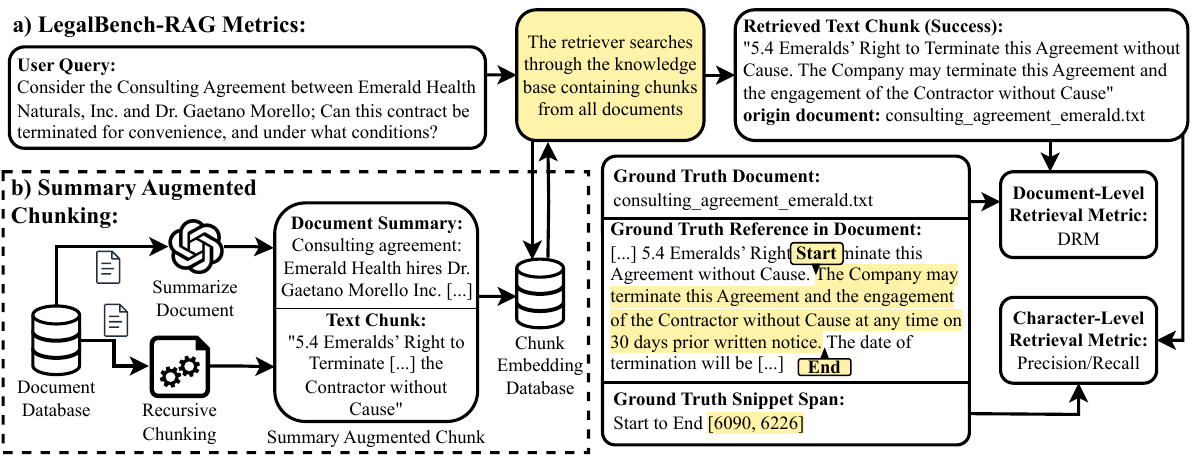}
    \caption{Part a) illustrates how our retrieval quality metrics, Document-Level Retrieval Mismatch (DRM) and text-level precision/recall, are computed in the LegalBench-RAG \cite{pipitone2024legalbench} information retrieval task. Part b) shows the process of setting up the knowledge base using Summary Augmented Chunks (SAC).}
    \label{fig:main}
\end{figure*}

\subsection{Focus on the Pre-Retrieval Stage: Chunking and Context Enrichment}

The performance of a RAG system is heavily influenced by the pre-retrieval phase, where the knowledge base documents are processed and indexed. The dominant practice is \emph{chunking}, that breaks down large documents into smaller pieces for efficient indexing in a vector database. This process must balance efficiency (smaller chunks), relevance (precise chunks), and context preservation (semantically complete chunks) \cite{barnett2024RAGfailurepoints, gao2023RAGsurvey}.
Naive chunking methods like fixed-size splitting can fragment logical units, leading to incomplete text snippets. More advanced strategies aim to preserve meaning. Recursive character splitting, for instance, divides text along natural boundaries like paragraphs and punctuation\footnote{\url{https://python.langchain.com/docs/how_to/recursive_text_splitter/}}. Semantic chunking uses language models to identify natural breakpoints, ensuring each chunk encapsulates a complete idea. Yet, even these methods struggle with legal text, as they may miss provisions spanning multiple sections or fail to handle nested clauses effectively \cite{ferraris2024legal, kalra2024hypa, qu2024semantic}.

This can be explained due to a fundamental limitation of chunking: the inevitable loss of global context. Each chunk is embedded as an isolated vector, disconnected from the broader document it belongs to. This isolation is a primary cause of what we identify and later formally define as Document-Level Retrieval Mismatch (DRM), a critical failure where the retriever selects chunks from entirely incorrect source documents that happen to share superficial similarities with the query. This is particularly problematic in legal databases with numerous structurally similar documents. While context loss is a known issue \cite{ferraris2024legal, gunther2024late}, DRM has not been formally quantified in the legal NLP literature.

To combat this context loss, various general context enrichment strategies have been developed. A straightforward local approach is to expand retrieved chunks to include surrounding sentences, a technique referred to as “Small2Big”. A more global approach involves adding metadata, which can be either standard (timestamps, authors, titles) or artificially generated. Examples of artificial metadata include chunk-specific explanatory context, such as in Contextual Retrieval by Anthropic\footnote{\url{https://www.anthropic.com/engineering/contextual-retrieval}}, or synthetic questions that a chunk could answer, as seen in methods like Reverse HyDE and QuIM-RAG \cite{gao2023precise, saha2024advancing}. Our Summary-Augmented Chunking falls into this last category, focusing on a lightweight, scalable approach where a single document-level summary provides global context to every chunk derived from it.

While our work focuses on this practical technique, other research has explored more architecturally complex solutions. These include methods that rethink the indexing structure, such as the hierarchical approach in RAPTOR \cite{sarthi2024raptor}, or knowledge graphs that model legal relationships \cite{kalra2024hypa}. The recent Late Chunking method \cite{gunther2024late} preserves more semantic context information by first embedding a document's full content and then performing chunking at the embedded level. Finally, the emergence of long-context models that can process hundreds of thousands of tokens presents a potential alternative to the chunking paradigm altogether, a use case specifically highlighted by OpenAI for lengthy legal documents \cite{OpenAICookbookLegalRAG}.

However, these advanced methods often introduce significant computational overhead or implementation complexity. Our research, in contrast, deliberately focuses on a practical, modular, and resource-efficient technique. Furthermore, we investigate a novel aspect of context enrichment by examining how domain expertise from legal professionals can be used to create more powerful, legally-informed summaries.

\section{Methodology}
\subsection{Task Definition and Dataset}
To evaluate improvements in a RAG pipeline, a benchmark must be able to isolate the performance of the retrieval stage from the final generative output. Widely adopted benchmarks like LegalBench \cite{guha2023legalbench} and LexGLUE \cite{chalkidis2021lexglue} are designed to test the intrinsic reasoning capabilities of LLMs. Consequently, when these are used to evaluate RAG systems, any performance changes are difficult to attribute specifically to the retrieval component, as its contribution is blended with the model's internal knowledge. For this reason, our study uses \emph{LegalBench-RAG} \cite{pipitone2024legalbench}, a recently developed benchmark specifically designed to isolate and evaluate the retrieval component of RAG systems in the legal domain. It is constructed from the well-established LegalBench corpus.

The LegalBench-RAG benchmark comprises multiple datasets that target distinct types of legal documents: \textbf{(i) CUAD} (Contract Understanding Atticus Dataset, \cite{hendrycks2021cuad}), which contains general contracts; \textbf{(ii) MAUD} (Merger Agreement Understanding Dataset, \cite{wang2023maud}), consisting of merger agreements; \textbf{(iii) ContractNLI} \cite{koreeda2021contractnli}, a dataset of non-disclosure agreements; and \textbf{(iv) PrivacyQA} \cite{ravichander2019question}, which includes privacy policies from mobile applications.

We measure the performance on LegalBench-RAG via document-level DRM and character-level precision/recall between the retrieved and ground-truth text snippets (see Figure \ref{fig:main}a), offering a holistic measure of retrieval quality. While our current work focuses exclusively on this retrieval analysis, we are currently working on adapting a benchmark such as the Australian Legal QA dataset\footnote{\url{https://huggingface.co/datasets/isaacus/open-australian-legal-qa}} for end-to-end performance evaluation in future work.

\subsection{Problem of Document-Level Retrieval Mismatch (DRM)}

We started by conducting diagnostic experiments with a standard RAG approach in LegalBench-RAG \cite{pipitone2024legalbench} to establish a baseline performance and understand the behavior of standard RAG systems on this task. We began by evaluating a range of retrieval architectures and embedding models to determine the typical performance ceiling and identify any systemic weaknesses. Across different configurations and models, overall retrieval scores remained consistently low (see Appendix \ref{app:model-ablation}).

We identified a major bottleneck: across different architectures and embedding models, retrievers frequently select chunks from entirely incorrect source documents. We define \textbf{Document-Level Retrieval Mismatch (DRM)} as the proportion of top-k retrieved chunks that do not originate from the document containing the ground-truth text.

\begin{figure*}[ht!]
\centering
\subfloat[DRM for Standard RAG]{\includegraphics[width=3in]{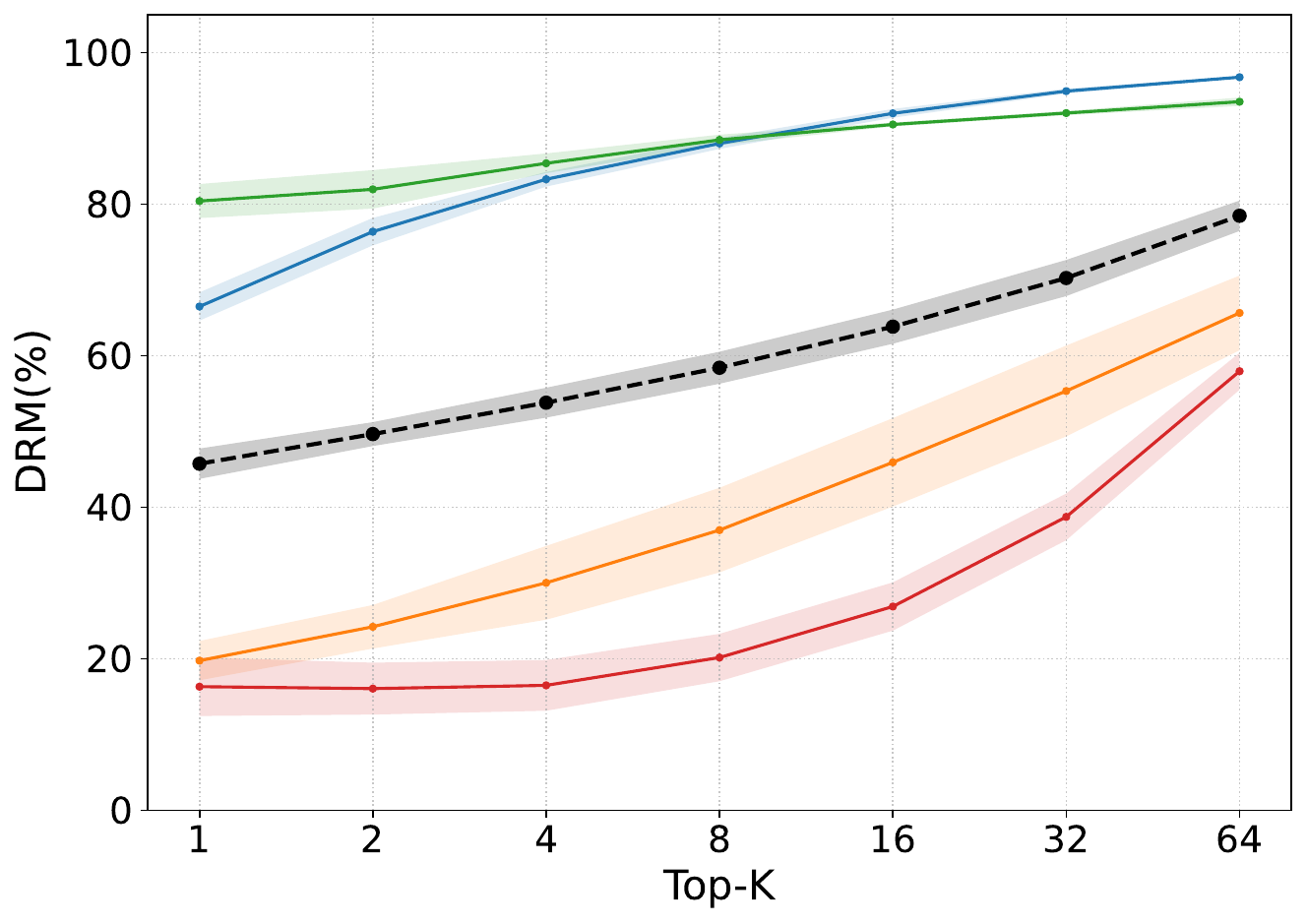}
\label{fig:ide_base}}
\hfil
\subfloat[DRM for RAG using SAC]{\includegraphics[width=3in]{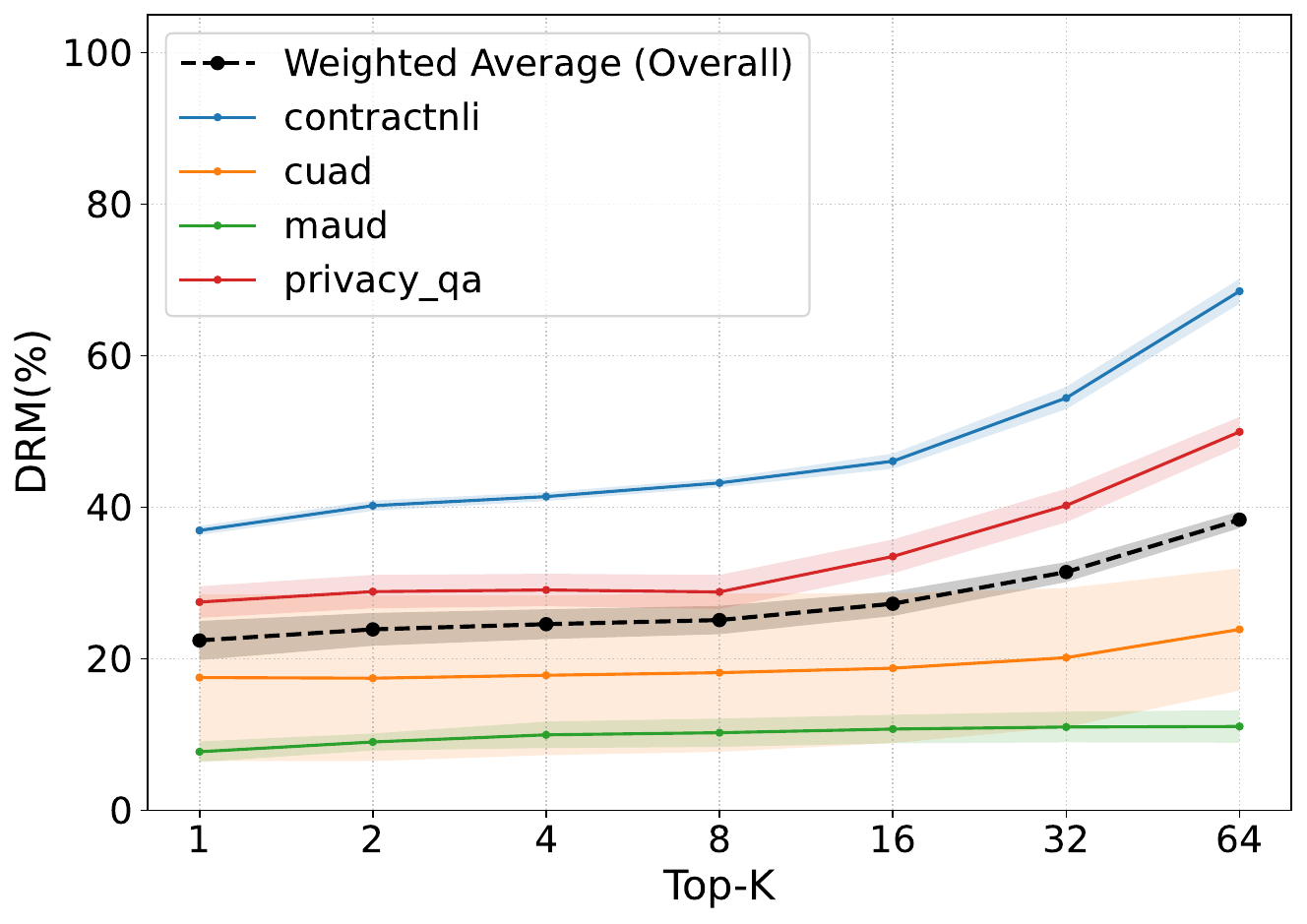}
\label{fig:ide_sac}}
\caption{Document-Level Retrieval Mismatch (DRM) of a standard RAG approach (left) and using our Summary Augmented Chunking (right), applied to the 4 datasets in the LegalBench-RAG benchmark. Retrieval using SAC selects fewer wrong documents across all top-k retrieved snippets and seeds.}
\label{fig:ide}
\end{figure*}

While DRM is a general challenge for retrieval systems and increases the probability of hallucinations in the subsequent generation of RAG systems \cite{hou2024gaps}, its impact is particularly severe in the legal domain due to the high degree of lexical and structural similarity across documents \cite{ferraris2024legal}. For instance, when we tested a standard RAG pipeline on ContractNLI data \cite{koreeda2021contractnli}, we observed DRM rates over 95\% (Fig. \ref{fig:ide_base}) in a pool of 362 documents. Our legal experts hypothesize that this may be due to the highly standardized, boilerplate nature of non-disclosure agreements, which are largely uniform apart from a few key variables. This linguistic homogeneity confuses retrieval models that rely on semantic similarity (or keyword matching), leading them to prioritize chunks that are textually similar to the query but from the wrong agreement. A concrete example of how the retriever fails on similar contracts can be found in Section \ref{sec:qualitative_results}.

This problem of lexical and structural similarity is not limited to contracts: in any legal task, users need assurance that retrieved context truly comes from the intended document. For example, when answering a question about a privacy policy, pulling text from a different but similar policy undermines both factual accuracy and trust in the system. Even if the generated answer happens to be correct, legal professionals expect document-faithful reasoning, making DRM a key measure of whether retrieval respects source boundaries.

\subsection{A Simple Solution: Summary Augmented Chunking (SAC)}

To combat DRM, we experimented with a simple methodology that we named Summary-Augmented Chunking (SAC). SAC works as follows (see Figure \ref{fig:main}b):

\textbf{(i) Summarization}: For each document in the corpus, we use an LLM to generate a single, concise summary as ``document fingerprint'', approximately 150 characters long. A detailed analysis of length and its impact is provided in Appendix \ref{app:size}.
\textbf{(ii) Chunking}: We employ a recursive character splitting strategy to partition the document's content into smaller, manageable chunks. This established method performs well on our dataset, as supported by prior work \cite{kalra2024hypa} and our own empirical results.
\textbf{(iii) Augmentation}: We prepend the document-level summary to each chunk derived from that document.
\textbf{(iv) Indexing}: The summary-augmented chunks are then embedded and indexed in a vector database for retrieval.

This approach injects crucial global context into each chunk, specifically to mitigate DRM by guiding the retriever to the correct source document. The method is highly practical, requiring only one additional LLM call per document and can be smoothly integrated into existing RAG pipelines with minimal computational overhead.
The generic prompt used for summarization is the following:

\begin{tcolorbox}[title=Generic Summarization Prompt, colback=white, colframe=black!40, breakable]
\footnotesize
\textbf{System:} You are an expert legal document summarizer.

\textbf{User:} Summarize the following legal document text. Focus on extracting the most important entities, core purpose, and key legal topics. The summary must be concise, maximum \{char\_length\} characters long, and optimized for providing context to smaller text chunks. Output only the summary text.

\textbf{Document:} \{document\_content\}
\end{tcolorbox}

Because LLMs often deviate from the specified length, we allowed a tolerance of 20 characters. Outputs exceeding this limit were regenerated with a reduced \texttt{char\_length} value.

\subsection{Expert-Guided Summarization}

While generic summaries provide a significant improvement, we hypothesize that tailoring summaries to the nuances of specific legal document types could further enhance retrieval performance. The especially high residual mismatch in datasets containing non-disclosure agreements and privacy policies (Fig. \ref{fig:ide_sac}) suggests that certain contractual language requires more sophisticated contextual cues, motivating our Expert-Guided summarization approach.

In collaboration with two legal experts\footnote{An associate professor and a postdoctoral researcher in law, with expertise in data protection and private law.}, we engineered a more sophisticated ``meta-prompt'' that instructs the LLM to generate summaries as \emph{distinct} as possible within a document type. It directs the model to identify and prioritize key differentiating legal variables. To test this, we focused on non-disclosure agreements and privacy policies, defining each type’s key characteristics from legally required elements and highlighting distinguishing features such as party names, definitions of data categories and their processing (for privacy policies), and definitions of confidential information (for NDAs). Our resulting expert-guided summarization prompt can be found in Appendix \ref{app:expert-prompt}.

This idea is supported by recent findings in legal NLP. For instance, research on summarizing Italian tax law decisions has demonstrated that a modular, expert-validated approach provides a solid basis for downstream semantic search, a task analogous to our retrieval application \cite{pisano2025summarization}. According to their findings, moving beyond generic summarization could be vital for complex legal texts. Their two-step method combines separate summary parts, whereas our ``meta-prompt'' uses conditional logic to integrate document type classification and summarization implicitly.

\begin{figure*}
\centering
\subfloat[Text-Level Precision]{\includegraphics[width=3in]{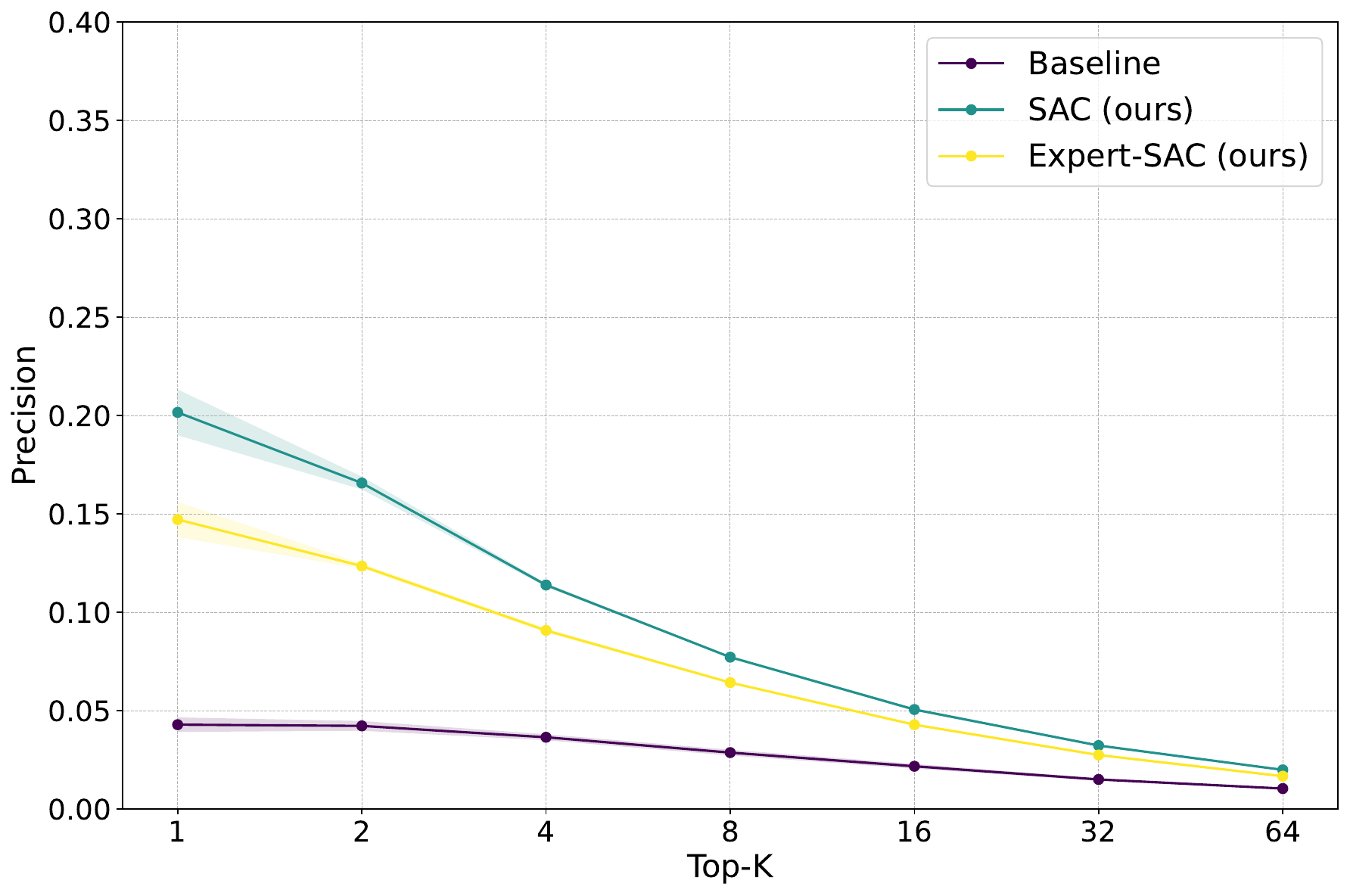}
\label{fig:precision}}
\hfil
\subfloat[Text-Level Recall]{\includegraphics[width=3in]{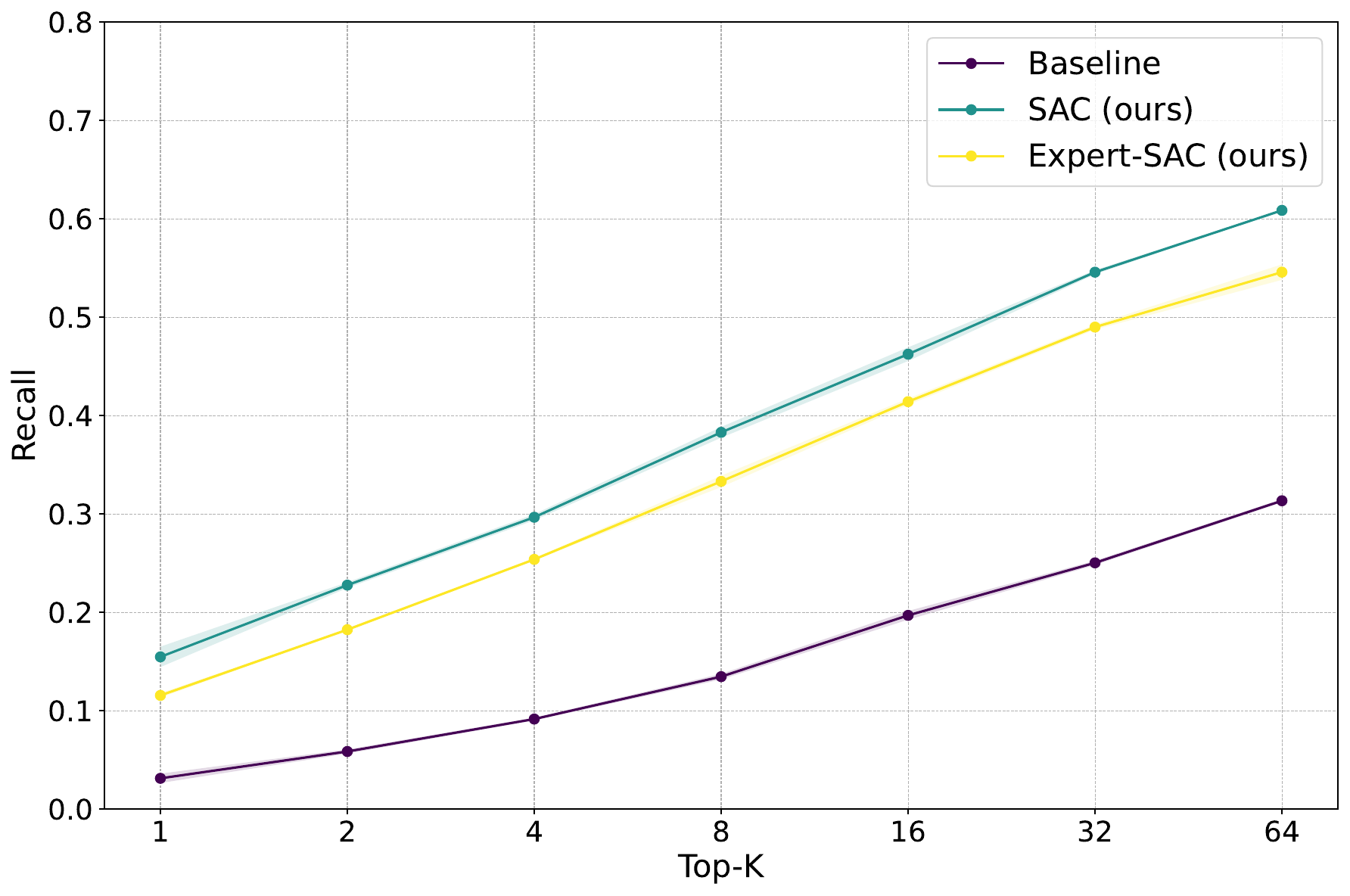}
\label{fig:recall}}
\caption{Text-level precision (left) and recall (right) of the standard RAG approach and SAC with general or expert-guided summarization strategy. The metrics are averaged over all datasets and seeds.}
\label{fig:precisionandrecall}
\end{figure*}

\section{Experimental Setup}

We evaluate the performance of our methods covering a broad picture of retrieval quality:

\textbf{(i) Document-Level Retrieval Mismatch (DRM)}: As our primary metric, DRM directly measures the retriever's ability to identify the correct source document. A lower DRM indicates higher precision at the document level.

\textbf{(ii) Text-Level Precision}: It measures the fraction of all the retrieved text that is part of the ground truth text span. High precision means that the retrieved context is concise and contains minimal irrelevant ``noise''.

\textbf{(iii) Text-Level Recall}: It evaluates what fraction of the ground truth text was found by the retrieval system. High recall indicates that the system found all the necessary information.

As our baseline, we implemented a standard RAG pipeline using a \emph{recursive character splitting} strategy with a chunk size of 500 characters and without overlap. The document summaries were generated using \texttt{gpt-4o-mini}~\cite{hurst2024gpt} and we instructed and processed the summaries to be about 150 characters long (details in the Appendix \ref{app:size}). The concatenated texts were embedded with \texttt{thenlper/gte-large}\footnote{\url{https://huggingface.co/thenlper/gte-large}} \cite{li2023towards} (see Appendix \ref{app:model-ablation}) and indexed in a FAISS \cite{johnson2019billion} vector database with cosine similarity as retrieval metric.

Dense semantic search excels at capturing conceptual similarity but may overlook exact lexical matches. In contrast, BM25 \cite{robertson2009probabilistic} is well established for keyword-based retrieval and can be effective for queries with unique identifiers or technical terminology. We therefore experimented with a hybrid dense+sparse retrieval. However, results showed that BM25 (sparse) improved DRM but decreased precision/recall while introducing additional computational overhead (see the Appendix \ref{app:dense-sparse}), so we decided to only use dense retrieval in the main experiments.

For all systems, we report performance across a range of top-$k$ retrieved chunks. This reflects real-world deployment scenarios, where practitioners must balance precision and recall depending on application needs. Reporting the full curve enables a more informative assessment of trade-offs across retrieval strategies.

\section{Results}
\subsection{Automatic Evaluation}

We demonstrate that SAC significantly reduces DRM compared to the baseline, showcasing its effectiveness in providing necessary global context. The results, reported in Figure \ref{fig:ide_sac}, show a dramatic reduction in DRM across a wide range of hyperparameters, effectively \emph{halving the mismatch rate}. Crucially, this improvement in document-level accuracy translates directly to improved text-level retrieval quality. By guiding the retriever to the correct document, RAG systems using SAC significantly outperform the standard RAG baseline on character-level precision and recall as well (Figure \ref{fig:precisionandrecall}).
Unexpectedly, Expert-Guided Summarization did not yield improvements over the general prompt (Figure \ref{fig:precisionandrecall}). It resulted only in slightly better retrieval metrics in a few specific settings, such as with larger chunk sizes.

\subsection{Qualitative Evaluation of Legal Experts}
\label{sec:qualitative_results}

Beyond quantitative metrics, a qualitative analysis offers critical insights into how different summarization strategies impact retrieval, especially where the baseline struggles. We illustrate these observations with a representative example from the ContractNLI dataset, focusing on a query about Non-Disclosure Agreements (NDAs), highlighting baseline failure and contrasting generic versus expert-guided SAC.
    
\begin{tcolorbox}[title=Example 1: NDA (contractnli), colback=white, colframe=black!40, breakable]
\footnotesize

\textbf{Query:} ``Consider Evelozcity's Non-Disclosure Agreement; Does the document allow the Receiving Party to independently develop information that is similar to the Confidential Information?''

\medskip

\textbf{Ground Truth:} \\
Relevant document: \emph{NDA-Evelozcity.txt}\\  
Relevant snippets:  
``The obligations of the Recipient specified in Section 2 above shall not apply with respect to Confidential Information to the extent that such Confidential Information:''  

``(d) is independently developed by or for the Recipient by persons who have had no access to or been informed of the existence or substance of the Confidential Information.''

\medskip
\textbf{A. Baseline Retrieval - No Summary} \\
Retrieved document: \emph{NDA-ROI-Corporation.txt} \ding{55} \\
Retrieved snippet:
``NON-DISCLOSURE AGREEMENT FOR PROSPECTIVE PURCHASERS'' \\
Comment: Complete failure, distracted by structural similarity of an other irrelevant NDA header.
\medskip

\textbf{B. Using a 150-character summary generated with the generic meta-prompt} \\
Retrieved document: \emph{NDA-Evelozcity.txt} \ding{52} \\
Summary: ``Non-Disclosure Agreement between Evelozcity and Recipient to protect confidential information shared during a meeting.'' \\
Retrieved snippet:
``; or (d) is independently developed by or for the Recipient by persons who have had no access to or been informed of the existence or substance of the Confidential Information.'' \\
Comment: Successful document-level retrieval (97\% precision, 50\% recall). The generic summary effectively guided to the correct document and relevant clause.

\medskip

\textbf{C. Using a 150-character summary generated with the expert meta-prompt} \\
Retrieved document: \emph{NDA-Evelozcity.txt} \ding{52} \\
Summary: ``NDA between Evelozcity and Recipient; covers vehicle prototypes, confidentiality obligations, exclusions, 5-yr term, CA governing law.'' \\
Retrieved snippet: 
``NON-DISCLOSURE AGREEMENT\\
This NON-DISCLOSURE AGREEMENT (this “Agreement”) is made as of this \_\_\_ day of \_\_\, 2019, by and between Evelozcity with offices at *address*
(the “Disclosing Party”), and \_\_ (the “Recipient”).'' \\
Comment: While the correct document was found, the snippet is the introductory boilerplate, completely useless for the query.

\medskip

\textbf{D. Using a 300-character summary generated with the expert meta-prompt} \\
Retrieved document: \emph{NDA-Evelozcity.txt} \ding{52} \\
Summary: ``**Definition of Confidential Information**: Non-public vehicle prototypes and company plans. **Parties**: Disclosing Party: Evelozcity, CA; Recipient: [Name Not Provided]. **Obligations**: Keep confidential, limit access to affiliates, use only for evaluation. **Exclusions**: Public knowledge, prior possession, independent development.'' \\
Retrieved snippet: The same snippet as in case C. \\
Comment: Similar to C, correct document, but irrelevant boilerplate snippet despite richer summary.

\end{tcolorbox}

The ground truth in Example 1 consists of two related snippets addressing the independent development of information similar to confidential material. It demonstrates the baseline's (A) complete failure to identify the correct document, highlighting the high Document-Level Retrieval Mismatch (DRM) caused by lexical redundancy and structural similarities in legal corpora. Both generic (B) and expert-guided (C, D) SAC approaches, however, successfully guided the retriever to the correct source document, clearly showing SAC's effectiveness in mitigating DRM by injecting global context.

Crucially, while document-level retrieval improved, a key difference emerged in snippet quality. The top-ranked chunk with the generic summary (B) was one of the correct snippets, directly engaging with the query. Conversely, both expert-guided summaries (C and D), despite retrieving the correct document, yielded an irrelevant introductory boilerplate snippet.

From a legal expert perspective, the expert-guided summaries (especially D) are richer, more structured and contain highly discriminative information for differentiating between NDAs (e.g., parties, subject matter, duration, exclusions). Yet, this legal assessment contrasts with observed retrieval performance. Expert summaries, while legally more informative and superior for differentiating documents, did not translate to better text-level snippet retrieval. This counter-intuitive result requires a more technical explanation (we explore in the next Section \ref{sec:discussion}) and suggests a complex interaction within the embedding space, where highly specific, dense legal information may not be optimally processed for general query alignment.

\subsection{Discussion}
\label{sec:discussion}
Our findings demonstrate that summary-based context enrichment provides a robust and scalable solution to a fundamental weakness of RAG in the legal domain: the loss of global context during chunking. By prepending document-level summaries to each text chunk, our method helps in guiding the retriever toward the correct source document, as evidenced by the drastic reduction in DRM. This is particularly valuable in legal corpora, where high structural similarity and standardized language make cross-document confusion a dominant failure mode. The success of this simple intervention underscores the importance of preserving document-level semantics in a domain where the overarching context dictates the meaning of individual clauses, a critical aspect for reliable legal NLP applications.

Interestingly, our experiments revealed that generic summaries consistently outperformed expert-guided ones, a counter-intuitive result given the legal precision of the latter. As highlighted in Section \ref{sec:qualitative_results}, from a legal perspective, expert-guided summaries successfully capture the distinctive elements required to differentiate between contracts of the same type. However, for the purpose of retrieval, we hypothesize two potential technical explanations for this observed performance gap.

First, generic summaries may strike a better balance between distinctiveness and broad semantic alignment with a wider variety of potential queries. While legally more precise, highly specific, expert-driven cues in the summaries might inadvertently overfit to narrow features. This would improve retrieval only in very specific cases and reduce robustness across a broader range of user intents.

Second, the informationally dense and structured language of expert-guided summaries may pose challenges for smaller embedding models, which must compress both the summary and chunk into a single vector. To investigate this potential bottleneck, future experiments with stronger, more capacious embedding models are needed.

In general, the interaction between summaries added to a chunk within the embedding space is complex. A strong global signal from the summary could overshadow the local relevance of a chunk. Understanding this dynamic is critical to improve our approach. Interesting insights from a machine learning perspective could be gained when analyzing the embedding space directly. We plan to use clustering and dimensionality reduction techniques to visualize how the concatenation of summaries and chunks behaves in the embedding space.

From a practical perspective, our results highlight the value of simple, modular interventions in the pre-retrieval stage. Unlike more complex architectural solutions (e.g., knowledge graphs, late chunking, or long-context models), SAC is inexpensive, requiring only a single additional summary per document, and integrates seamlessly into existing RAG pipelines. This makes it scalable even to large and dynamically changing legal databases. For practitioners, generic SAC provides an easily adoptable technique that delivers tangible improvements without the need for domain-specific fine-tuning or significant infrastructure changes.

Finally, our findings contribute to the broader vision of “LLMs as legal readers” \cite{palka2025make}. If future legal documents become longer and more comprehensive, retrieval reliability will be even more critical. Our approach represents a practical step toward building systems that can process such documents with greater reliability, making AI a more trustworthy partner in navigating the complexity of legal texts.

\subsection{Limitations and Future Work}
While promising, our work has several limitations. First, our experiments were restricted to particular categories of legal documents and conducted exclusively in English. These documents, while diverse, do not cover the full spectrum of legal text types, such as legislation, case law, or other types of contracts, which differ substantially in structure and interpretation. Moreover, legal meaning is highly jurisdiction-specific, and our datasets were largely restricted to common-law contexts.

Second, our analysis focused on an isolated intervention within a standard RAG pipeline to clearly measure its impact. While effective, the residual retrieval mismatch rates remain significant, indicating that SAC is a valuable component for reliable RAG but not a complete solution on its own. We believe that combining SAC with other well-researched modules is the most promising path toward achieving the reliability required for legal applications, with the next promising steps being:

\textbf{(i)} Extending the presented principle of summarization hierarchically, with summaries at the paragraph, section, and document level to provide context at multiple granularities.  
\textbf{(ii)} Applying query optimization methods (e.g., transformation, expansion, or routing) to bridge the semantic gap between user questions and the formal language of legal text chunks. 
\textbf{(iii)} Adding a reranking step where a more powerful model re-evaluates and re-orders the top-k retrieved chunks to improve the final selection before generation.
It would also be valuable to benchmark SAC against other context-preserving chunking strategies, such as Late Chunking \cite{gunther2024late} and RAPTOR \cite{sarthi2024raptor}, to better understand its relative strengths.

Finally, this study focuses exclusively on the retrieval stage of the RAG framework. Future work will investigate how the DRM metric and SAC impact downstream generation through end-to-end benchmarking.

\section{Conclusion} 
We addressed the critical challenge of retrieval reliability in RAG systems operating on large, structurally similar legal document databases. We identified and quantified \emph{Document-Level Retrieval Mismatch} (DRM) as a dominant failure mode, where retrievers are often easily confused by legal boilerplate language and select text from entirely incorrect documents. Targeting this issue, we investigate \emph{Summary-Augmented Chunking} (SAC), a simple and computationally efficient technique that prepends document-level summaries to each text chunk. By injecting global context, SAC drastically reduces DRM and consequently improves text-level retrieval precision and recall.

A key finding is that generic summaries outperform expert-guided ones focusing on key legal variables. For the purpose of guiding retrievers, broad semantic cues appear more robust and generalizable than dense, structured, legally precise summaries. This demonstrates that meaningful retrieval performance gains are achievable without heavy domain-specific engineering.

While SAC is not a full solution on its own, it offers a practical, scalable intervention for more reliable legal RAG systems. By improving the crucial retrieval step for legal information, our work brings us closer to a future where AI can truly serve as a trusted partner in the legal profession.

\section*{Acknowledgments}
This work is funded by the European Union. Views and opinions expressed are, however, those of the author(s) only and do not necessarily reflect those of the European Union or the European Health and Digital Executive Agency (HaDEA). Neither the European Union nor the granting authority can be held responsible for them. Grant Agreement no. 101120763 - TANGO.
This work was also partially supported by the 
CompuLaw – Computable Law – funded by the ERC under the Horizon 2020 (Grant Agreement N. 833647).


\bibliography{main}

\appendix

\section{Hyperparameter Chunk Size and Summary Size}
\label{app:size}
We experimented with chunk sizes of 200, 500, and 800 characters, combined with prepended summaries of either 150 or 300 characters. The precision and recall results for all six configurations are reported in Table \ref{tab:chunk-ablation}. For our final pipeline, we selected a chunk size of 500 characters, consistent with \citet{pipitone2024legalbench}, and a 150-character summary, as this configuration yielded the most balanced trade-off between precision and recall.

\begin{table}[ht]
\centering
\begin{tabularx}{\linewidth}{@{}X P{.65cm}P{.65cm}P{.65cm}P{.65cm}P{.65cm}P{.65cm}P{0.22cm}@{}}
\toprule
Chunk & \multicolumn{2}{c}{200} & \multicolumn{2}{c}{500} & \multicolumn{2}{c}{800} \\ 
\cmidrule(l{0.5em}r{0.5em}){2-3} \cmidrule(l{0.5em}r{0.5em}){4-5} \cmidrule(l{0.5em}r{0.5em}){6-7}
Sum. & 150 & 300 & 150 & 300 & 150 & 300 \\
\midrule
Prec.(\%) & 10.64 & 8.05 & \textbf{11.03} & 8.45 & 7.76 & 6.79 \\
Rec.(\%)  & 23.43 & 22.91 & 41.80 & 37.77 & 42.93 & \textbf{43.90} \\
DRM(\%)  & 20.70 & 23.49 & \textbf{19.29} & 20.61 & 34.96 & 29.68 \\
\bottomrule
\end{tabularx}
\caption{Average Document-Level Retrieval Mismatch (DRM), precision, and recall across different chunk sizes (\textit{Chunk}) and summary lengths (\textit{Sum.}), both measured in characters. Reported values are averaged over seven top-$k$ retrieval settings ($k \in {1,2,4,8,15,32,64}$). Lower values indicate better performance for DRM, while higher values are better for precision and recall.}
\label{tab:chunk-ablation}
\end{table}

\section{Dense Semantic Search or Sparse Keyword Search?}
\label{app:dense-sparse}
BM25 \cite{robertson2009probabilistic} is a well-established keyword-based (sparse) retrieval method. A common assumption in RAG research is that hybrid approaches, combining dense vector similarity with keyword matching, often yield the best results \cite{Berntson2023AzureAI}. Following this intuition, we augmented our dense retriever with a BM25 component, allowing the system to explicitly match salient terms (e.g., party names) from the prepended summaries with the same terms in the user query. Results are reported in Table \ref{tab:bm25-ablation}.

As expected, adding sparse retrieval improved document selection, reducing our document-level mismatch (DRM). However, it lowered text-level precision and recall. A closer inspection suggests a likely explanation: The summaries are highly structured and rich in identifiers, where sparse keyword matching excels, whereas the chunked document bodies contain more natural language and nearly no direct keywords to match. For pinpointing the relevant passage, semantic similarity between the query and text is more informative than keyword overlap, explaining the higher text-level precision and recall without the BM25 component.

In conclusion, adding BM25 search only contributed slightly to finding the correct document (improved DRM), but tended to result in poorer search results within a document (reduced precision and recall). Also considering the computational overhead of the BM25 algorithm, we decided to rely exclusively on semantic search.

\begin{table}[ht]
\centering
\begin{tabularx}{\linewidth}{@{}X P{1.0cm}P{1.0cm}P{1.0cm}P{1.0cm}P{1.3cm}@{}}
\toprule
$w_{semantic}$ & 100\% & 75\% & 50\% & 25\% \\ 
$w_{keyword}$ & 0\% & 25\% & 50\% & 75\% \\
\midrule
Prec.($\uparrow$\%) & \textbf{11.03} & 10.57 & 9.54 & 8.23 \\
Rec.($\uparrow$\%) & 41.80 & \textbf{42.54} & 41.47 & 36.56 \\
DRM($\downarrow$\%) & 19.29 & 19.11 & 18.45 & \textbf{18.18} \\
\bottomrule
\end{tabularx}
\caption{Average Document-Level Retrieval Mismatch (DRM), precision, and recall for different weightings of semantic similarity ($w_{semantic}$) and lexical similarity ($w_{keyword}$, BM25). Metrics are averaged over seven top-$k$ values ($k \in \{1,2,4,8,15,32,64\}$). Lower DRM and higher precision/recall indicate better performance. Results are based on the optimal chunk size of 500 characters and summary length of 150 characters identified in Section~\ref{app:size}.}
\label{tab:bm25-ablation}
\end{table}

\section{Embedding Model Ablation}
\label{app:model-ablation}
\begin{figure*}
\centering
\subfloat[Text-Level Precision]{\includegraphics[width=3in]{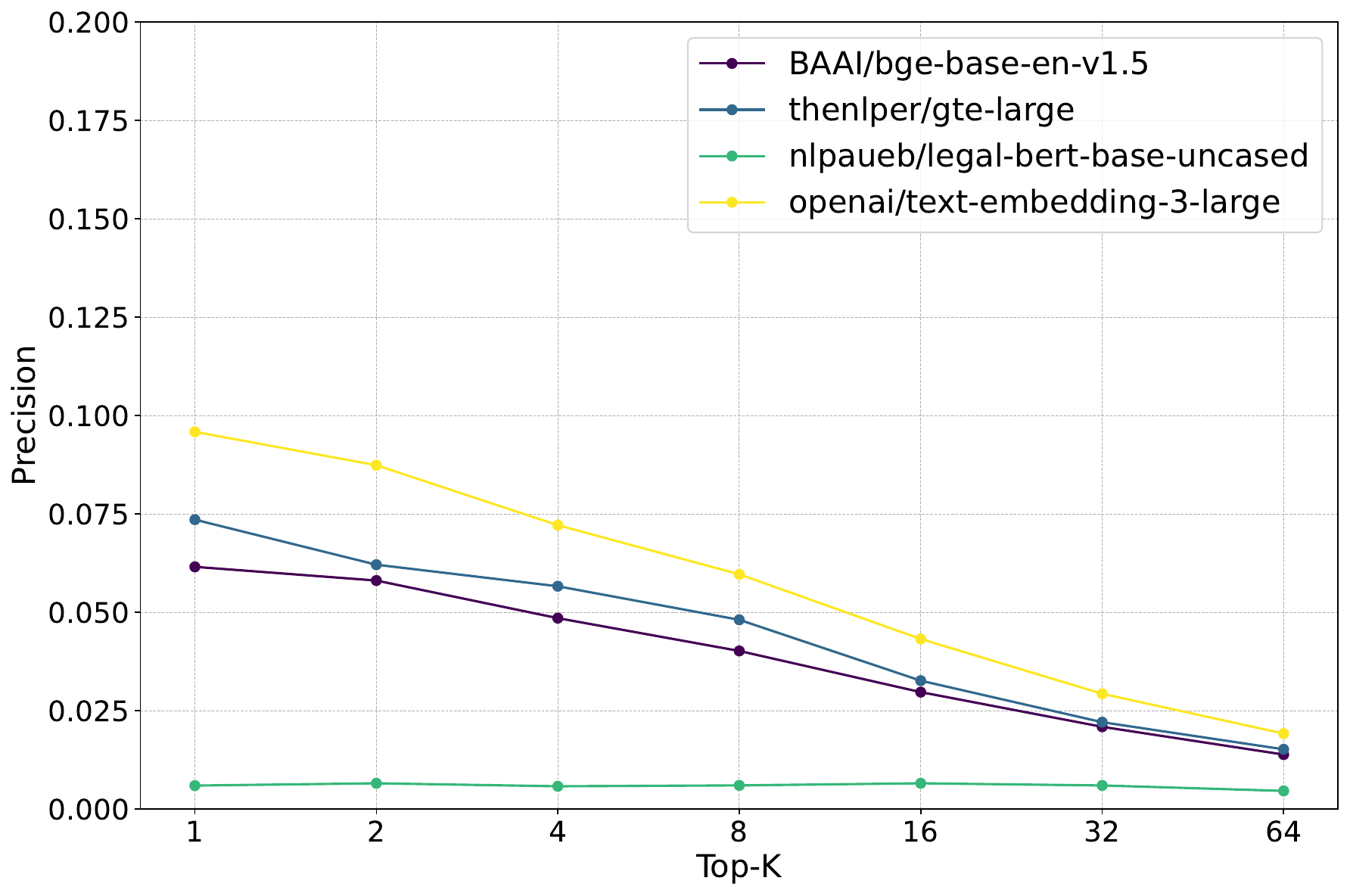}
\label{fig:embd-precision}}
\hfil
\subfloat[Text-Level Recall]{\includegraphics[width=3in]{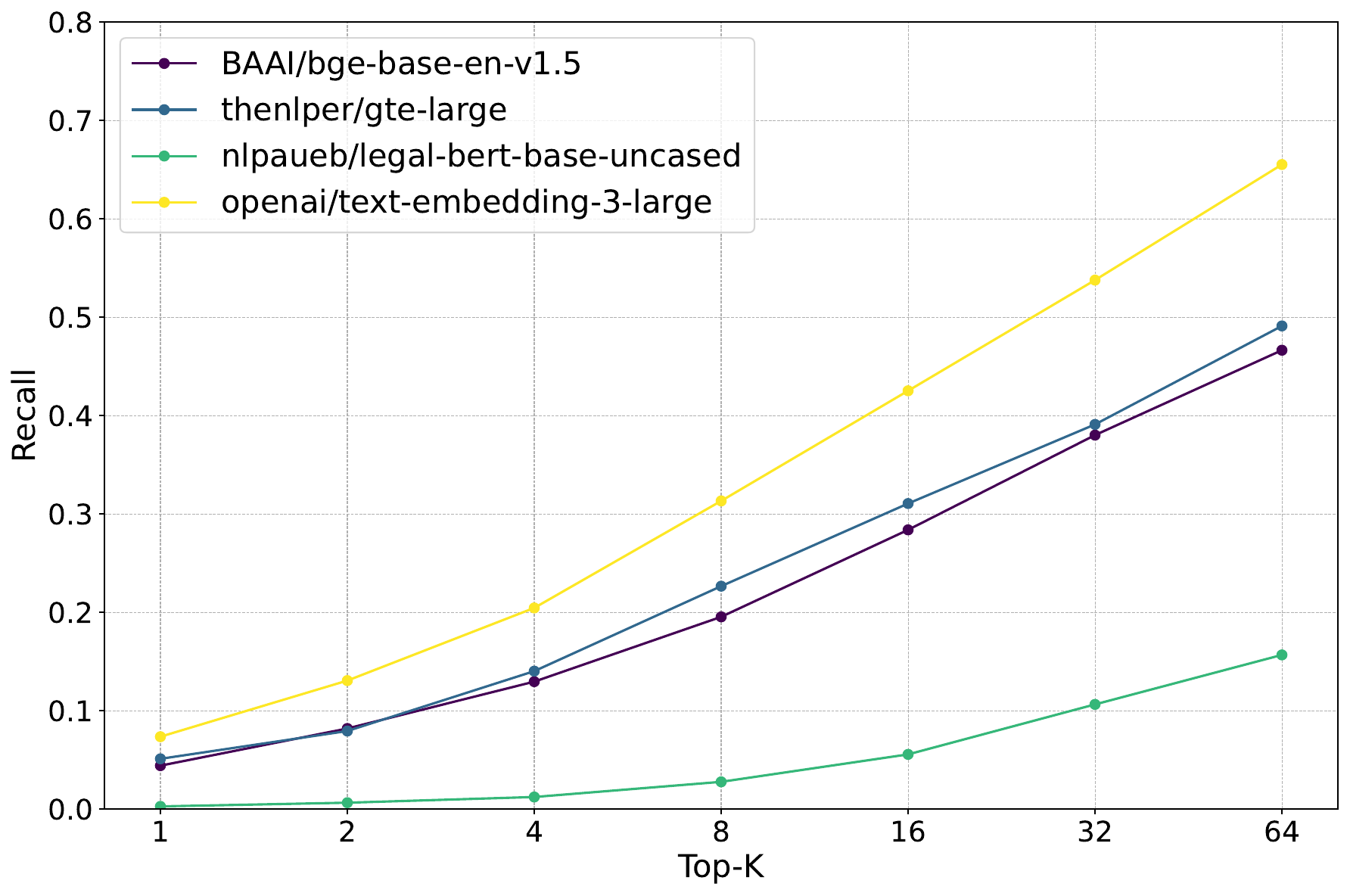}
\label{fig:embd-recall}}
\caption{Relative performance comparison of four embedding models in the baseline case on the LegalBench-RAG dataset \cite{pipitone2024legalbench}.}
\label{fig:embed-ablation}
\end{figure*}

The choice of embedding model plays a crucial role in retrieval performance, as embeddings determine how effectively the system identifies text snippets semantically similar to a given query. To assess this impact, we conducted a brief model ablation study, with results shown in Figure~\ref{fig:embed-ablation}. Among the tested models, OpenAI's \texttt{text-embedding-3-large}\footnote{\url{https://openai.com}} achieved the strongest results overall. However, due to concerns about API rate limits and the importance of reproducibility, we opted for an open-source alternative. Within this category, the best-performing model was \texttt{thenlper/gte-large}\footnote{\url{https://huggingface.co/thenlper/gte-large}}, which we therefore selected for all subsequent experiments.  

\section{Expert-Guided Prompt Template}
\label{app:expert-prompt}
Together with legal experts, we developed an expert-informed prompt template aimed at generating more distinctive legal text summaries. We focused on two document types: Non-Disclosure Agreements and privacy policies, that posed particular challenges for the generic prompt. The LLM was instructed to first identify the document type and then apply the corresponding template for summarization. The full prompt used in our experiments is provided below.

\begin{tcolorbox}[title=Expert-Guided Prompt Template, colback=white, colframe=black!40, breakable]
\footnotesize
\textbf{System:} You are a legal summarization expert.

\textbf{User:} Your task is to generate a highly distinct, structured summary of the provided legal document. The primary goal is to extract the unique identifiers that differentiate this document from others of the same type. This summary will be used as context to smaller text chunks for a retrieval system.

Follow this two-step process:

- First, internally identify the document type from the following options: Non-Disclosure Agreement (NDA), Privacy Policy, or Other.

- Second, generate the summary based on the specific template corresponding to the identified document type.

\medskip

\textbf{Document type Non-Disclosure Agreement (NDA):} An NDA is a legally binding contract between specific parties that outlines confidential information to be kept secret.
If the document is an NDA, your summary should align with the following template:

- Definition of Confidential Information, specifying what types of information are considered confidential, e.g. such as: Technical data, Business plans, Customer lists, Trade secrets, Financial information

- Parties to the Agreement identifying the disclosing party and the receiving party (or both, if mutual NDA), e.g. such as: Full legal names, Affiliates or representatives covered, Roles of each party

- Obligations of the receiving party outlining what the receiving party is required to do, e.g. such as: Keeping the information secret, Limiting disclosure to authorized personnel, Using the information only for specified purposes

- Exclusions from confidentiality describing information that is not protected under the NDA, such as: Information already known to the receiving party, Publicly available information, Information disclosed by third parties lawfully, Independently developed information

- Specifying any exceptions where disclosure is allowed, such as: To employees or advisors under similar obligations, If required by law or court order (with notice to the disclosing party)

- Term and Duration, defining how long the confidentiality obligation lasts: Often includes both the duration of the agreement and the period during which information remains protected (e.g., “3 years after termination”)

- Purpose of Disclosure (Use Limitation), stating the specific reason the information is being shared (e.g., for evaluating a partnership, conducting due diligence, etc.) and prohibits other uses.

- Remedies for Breach, detailing the consequences of violating the NDA, which may include: Injunctive relief (court orders to stop disclosure), Damages, Legal fees

- Governing Law and Jurisdiction, identifying which country/state’s laws apply and where disputes will be settled.

- Miscellaneous Clauses (Boilerplate), may include: No license granted, Entire agreement clause, Amendment process, Counterparts and signatures

\medskip

\textbf{Document type Privacy Policy:} A privacy policy is issued by a private or public entity to inform users how their personal data is processed (e.g., collected, used, shared, stored).
If the document is a privacy policy, your summary should align with the following template:

- Personal Data Collected and Processed, specifying what categories of personal data are collected and how. This may include: Name and surname, Contact information, Financial details, Device and browser data, Location information, Inferred preferences or behaviors

- Identity and Contact Details of the Controller, identifying the organisation responsible for the processing. May include: Full legal name of the controller, Contact email or phone number, Details of any representative (if applicable)

- Purposes of Processing, outlining why the personal data is collected and how it will be used. Examples include: Service provision and operation, Personalisation of content or features, Marketing and advertising, Analytics and performance monitoring, Payment processing

- Legal Basis for Processing, specifying the lawful grounds relied upon. These are: Consent of the data subject, Performance of a contract, Compliance with a legal obligation, Protection of vital interests, Task carried out in the public interest, Legitimate interests of the controller or third party

- Recipients of the Data, listing who may receive the data, including: Service providers and processors, Business partners, Public authorities (where legally required), Affiliates and subsidiaries

- International Data Transfers, describing whether personal data is transferred outside the jurisdiction and, if so: Destination countries, Safeguards applied (e.g., Standard Contractual Clauses, adequacy decisions)

- Data Retention, defining how long the personal data will be stored, or the criteria for determining the period. May include: Fixed retention periods, Purpose-based retention (e.g., “as long as necessary to provide the service”), Archiving or deletion policies

- Data Subject Rights, explaining individuals’ rights under data protection law, including: Right to access personal data, Right to rectify inaccuracies, Right to erasure (“right to be forgotten”), Right to restrict or object to processing, Right to data portability

- Right to Lodge a Complaint, providing information on: The data subject’s right to contact a supervisory authority, Name or link to the competent authority

- Automated Decision-Making, disclosing whether such processing occurs and, if so: The logic involved, Potential significance of the decisions, Expected consequences for the data subject

\medskip

\textbf{Other document type:} If the document does not match the types above, summarize the following general legal document in a structured, concise way. Identify for your summary the important entities, core purpose, and other unique identifiers that differentiate this document from others of the same type.

\medskip

\textbf{General Rules:}

- The summary must be concise and under \{char\_length\} characters.

- Ignore every field in the template where the information is not present in the document.

- Prioritize extracting the most critical identifiers, such as parties, dates, and the specific subject matter.

- Output ONLY the final summary text!

Here is the document you should summarize:
\{document\_content\}
\end{tcolorbox}

\end{document}